%% file: main.tex
\documentclass[lettersize,journal]{style/IEEEtran}

\PassOptionsToPackage{numbers, compress}{natbib}

\usepackage[utf8]{inputenc} 
\usepackage[T1]{fontenc}    
\usepackage{hyperref}       
\usepackage{url}            
\usepackage{booktabs}       
\usepackage{amsfonts}       
\usepackage{nicefrac}       
\usepackage{microtype}      
\usepackage[table]{xcolor}
\usepackage{amsmath}
\usepackage{amssymb}
\usepackage{mathtools}
\usepackage{amsthm}
\usepackage{microtype}
\usepackage{graphicx}
\usepackage{subfigure}
\usepackage{booktabs} 
\usepackage{hyperref}
\usepackage[none]{hyphenat}
\usepackage{makecell}

\title{Leveraging Continuously Differentiable Activation for Learning in Analog and Quantized Noisy Environments}
\author{%
  Vivswan Shah and Nathan Youngblood
  \thanks{This work was supported in part by the U.S. National Science Foundation under Grants CISE-2105972 and ECCS-2337674 and by AFOSR under Grant FA9550-24-1-0064.
  
  This research was supported in part by the University of Pittsburgh Center for Research Computing, RRID:SCR\_022735, through the resources provided. Specifically, this work used the H2P cluster, which is supported by NSF award number OAC-2117681.
  
  V. Shah and N. Youngblood are with the Department of Electrical and Computer Engineering, Swanson School of Engineering, University of Pittsburgh, Pittsburgh, PA 15261 USA (email: vivswanshah@pitt.edu and nathan.youngblood@pitt.edu).
  
   Code available at: \href{https://github.com/Vivswan/GeLUReLUInterpolation}{https://github.com/Vivswan/GeLUReLUInterpolation}.}
}

\begin{document}
\maketitle
\input{sections/0-abstract}
\input{sections/1-Introduction}
\begin{figure*}[t]
  \includegraphics[width=\textwidth]{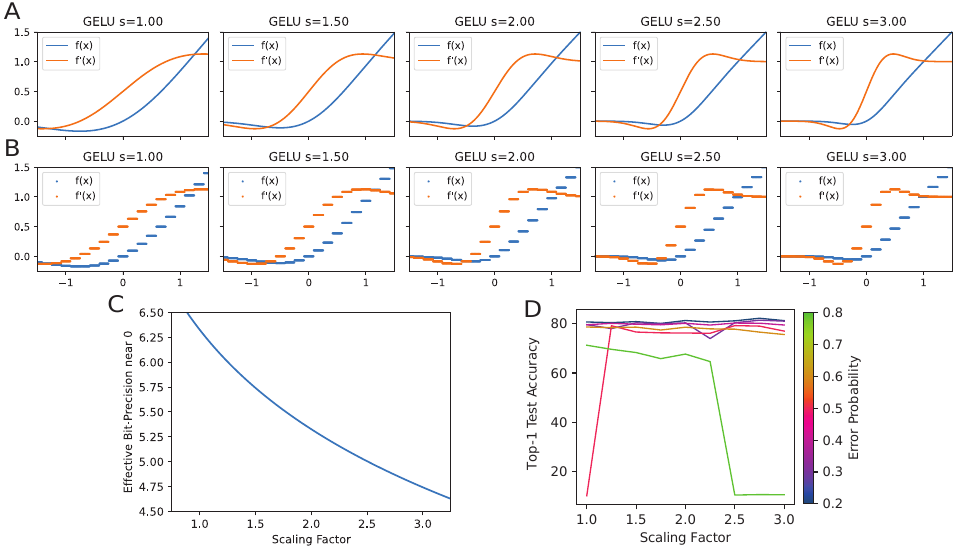}
  \caption{\textbf{Effects of Scaling Factor in GELU.} GELU function and its derivative at different values of scaling factor with \textbf{a)} with full precision; \textbf{b)} with reduced precision. \textbf{c)} The effective bit-precision of GELU derivative near zero at different values of scaling factor when input precision is set to 6-bits. \textbf{d)} Top-1 test accuracy of ConvNet marginally declines with increasing GELU scaling factor on CIFAR-10. }
  \label{fig:FunctionalAnalysis1}
\end{figure*}
\input{sections/2-Background}
\begin{figure*}[t]
  \includegraphics[width=\textwidth]{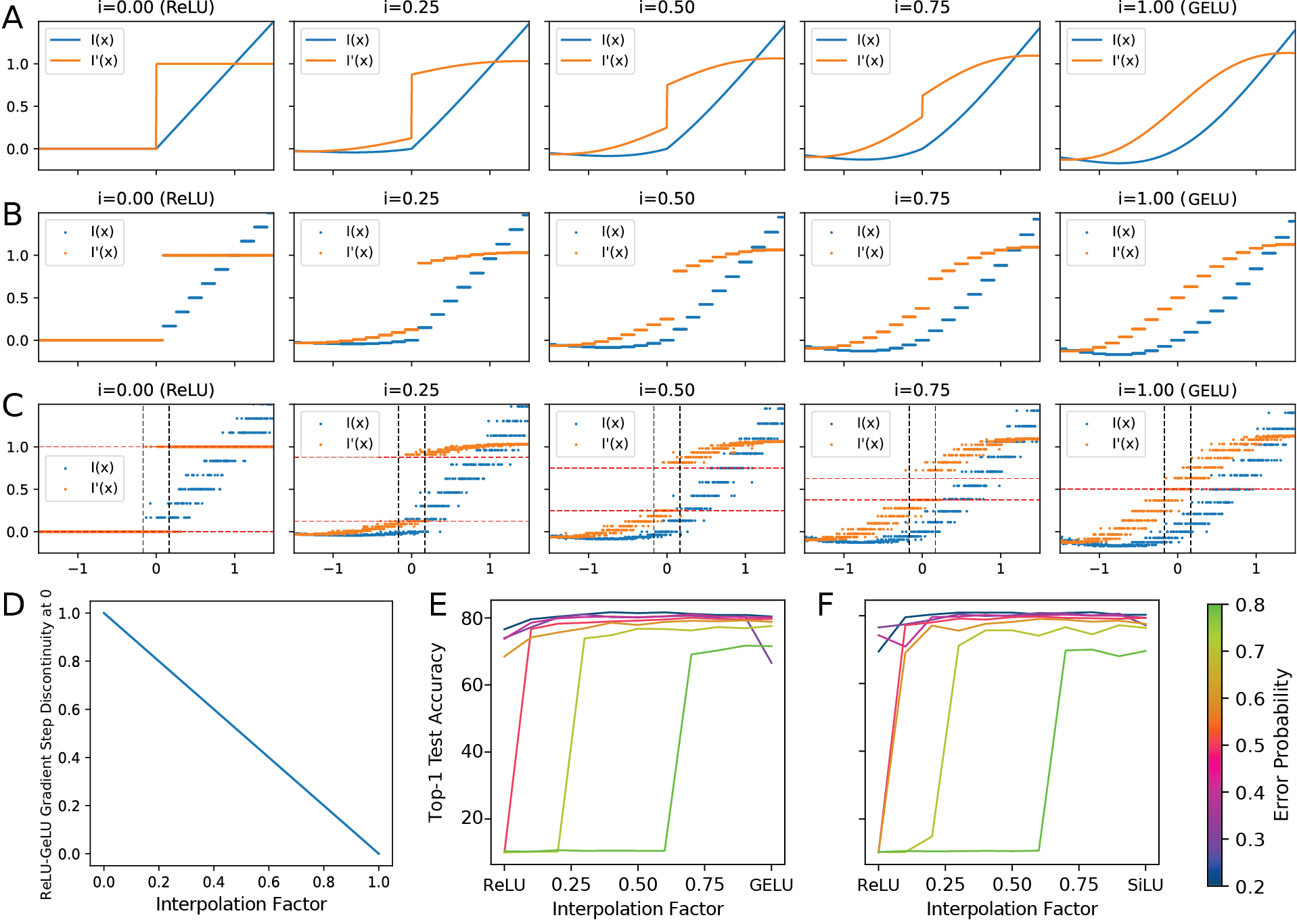}
  \caption{\textbf{Interpolation Factors.} ReLU-GELU interpolation function and its derivative at different values of interpolation factor:  \textbf{a)} at full-precision \textbf{b)} with reduced precision; \textbf{c)} with reduced precision and noise. \textbf{d)} The Gradient Step Discontinuity in ReLU-GELU interpolation at zero is negatively correlated to interpolation factor. \textbf{e) \& f)} Top-1 Test Accuracy of ConvNet Utilizing Linear Interpolation Activation Functions, \textbf{(e)} ReLU-GELU and \textbf{(f)} ReLU-SiLU, with Quantized Noise on CIFAR-10 Dataset.}
  \label{fig:ExperimentalResults1}
\end{figure*}
\input{sections/3-FunctionalAnalysis}

\begin{figure*}[t]
  \includegraphics[width=\textwidth]{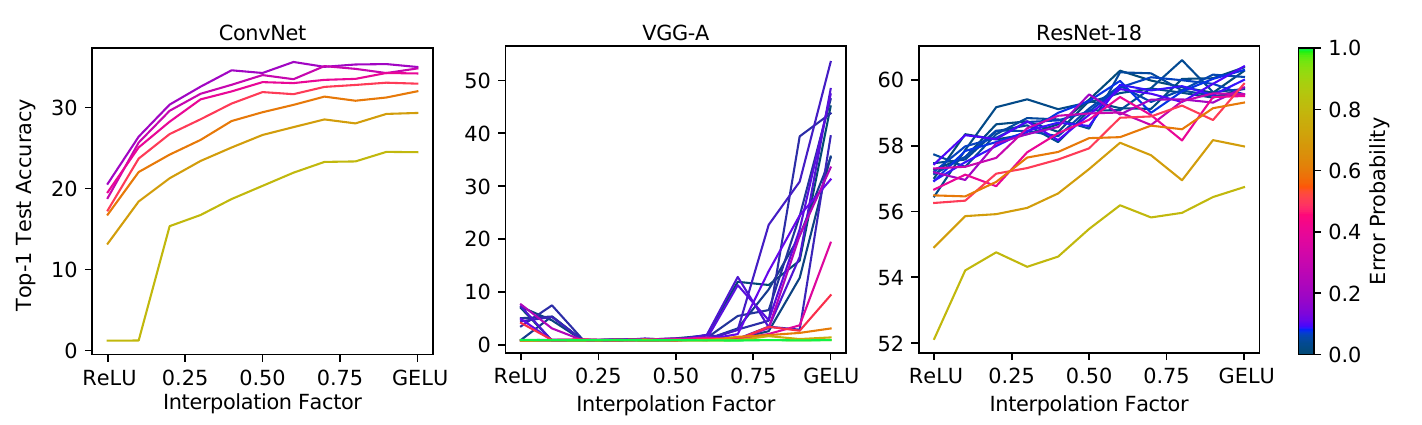}
  \caption{\textbf{Models evaluated on the CIFAR-100 dataset.} ReLU-GELU interpolation function and its derivative at different values of interpolation factor for ConvNet, VGG-A and ResNet-18 models on CIFAR-100}
  \label{fig:ExperimentalResultsCIFAR100}
\end{figure*}
\input{sections/4-ErrorAnalysis}
\begin{figure*}[t]
  \includegraphics[width=\textwidth]{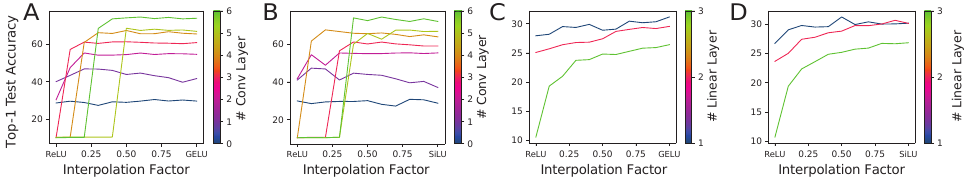}
  \caption{\textbf{Impact of Layer Depth in ConvNet Architecture on CIFAR-10 Dataset.} For (a) and (b) the number of convolutional layers is varied while only one linear layer is used. For (c) and (d) the number of linear layers is varied with no convolutional layers. In (a) and (c) ReLU-GELU interpolation is used, while in (b) and (d) ReLU-SiLU interpolation is used.}
  \label{fig:ExperimentalResults2}
\end{figure*}
\input{sections/5-Results}
\begin{figure*}[t]
  \includegraphics[width=\textwidth]{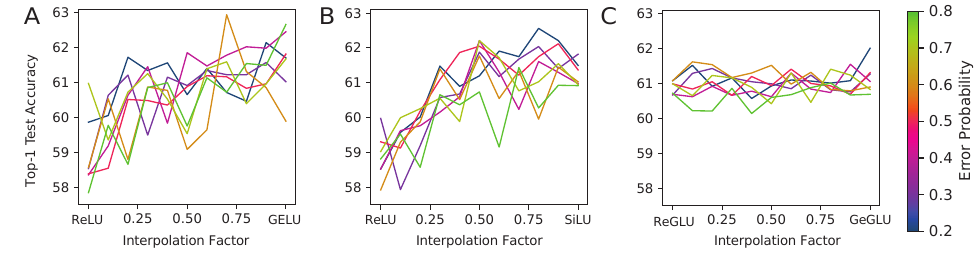}
  \caption{\textbf{Top-1 test accuracy of Vision Transformer (ViT) with analog photodetector/sensor on CIFAR-10 Dataset.}. The test accuracy results of the Vision Transformer when using (a) ReLU-GELU interpolation, (b) ReLU-SiLU interpolation, and (c) ReGLU-GeGLU interpolation}
  \label{fig:ExperimentalResults3}
\end{figure*}
\input{sections/6-Discussion}
\input{sections/7-Conclusion}

\bibliographystyle{IEEEtran}
\bibliography{IEEEabrv, references}

\newpage
\input{sections/10-Appendix}

\end{document}

%% file: sections/0-abstract.tex
\begin{abstract}
Real-world analog systems, such as photonic neural networks, intrinsically suffer from noise that can impede model convergence and accuracy for a variety of deep learning models. In the presence of noise, some activation functions behave erratically or even amplify the noise. Specifically, ReLU, an activation function used ubiquitously in digital deep learning systems, not only poses a challenge to implement in analog hardware but has also been shown to perform worse than continuously differentiable activation functions. In this paper, we demonstrate that GELU and SiLU enable robust propagation of gradients in analog hardware because they are continuously differentiable functions. To analyze this cause of activation differences in the presence of noise, we used functional interpolation between ReLU and GELU/SiLU to perform analysis and training of convolutional, linear, and transformer networks on simulated analog hardware with different interpolated activation functions. We find that in ReLU, errors in the gradient due to noise are amplified during backpropagation, leading to a significant reduction in model performance. However, we observe that error amplification decreases as we move toward GELU/SiLU, until it is non-existent at GELU/SiLU demonstrating that continuously differentiable activation functions are $\sim$100$\times$ more noise-resistant than conventional rectified activations for inputs near zero. Our findings provide guidance in selecting the appropriate activations to realize reliable and performant photonic and other analog hardware accelerators in several domains of machine learning, such as computer vision, signal processing, and beyond.
\end{abstract}

%% file: sections/1-Introduction.tex
\section{Introduction}
Rapid advancement of artificial intelligence and deep learning has sparked interest in novel computing paradigms that can overcome the limitations of traditional digital electronics \cite{mehonic_brain-inspired_2022}. Photonic neural networks have emerged as a promising approach, offering the potential for ultra-fast, energy-efficient computation by leveraging the properties of light \cite{shastri_photonics_2021}. However, the transition from digital to analog photonic systems introduces new challenges, particularly in handling noise and maintaining computational accuracy.

Unlike their digital counterparts, photonic neural networks are implemented in analog hardware like coherent \cite{lin_all-optical_2018, shen_deep_2017, kari_integrated_2024, Computational, youngblood_computational_2023, mourgias-alexandris_neuromorphic_2020, youngblood_realization_2023, chen_deep_2023, youngblood_coherent_2022, rahimi_kari_realization_2024}, electro-absorptive \cite{giamougiannis_silicon-integrated_2021}, phase-change \cite{feldmann_parallel_2020, erickson_designing_2022}, magneto-optics \cite{pintus_integrated_2024, murai_nonvolatile_2020, youngblood_non-reciprocal_2024-1}, microring resonator \cite{tait_broadcast_2014}, and dispersive fiber-based architectures \cite{xu_11_2021}. Due to their physical and analog nature, signals in these devices are continuous and subject to various sources of noise. This includes shot noise, thermal noise, and quantization errors in optical-to-electrical conversions. In this context, the choice of activation function becomes crucial, as it significantly impacts the network's ability to learn and generalize in the presence of noise.

Traditionally, rectified linear units (ReLU) \cite{nair_rectified_2010} have been widely used in digital neural networks due to their simplicity and effectiveness in mitigating the vanishing gradient problem \cite{pascanu_difficulty_2013}. However, ReLU's discontinuous nature at zero can lead to instabilities in gradient propagation when implemented in analog photonic systems. This discontinuity can amplify noise effects, potentially degrading the network's performance and reliability. Thus, attempts to directly mimic activation functions optimized for digital neural networks, such as ReLU, can actually be counterproductive for analog accelerators.

To address these challenges, we propose the use of continuously differentiable activation functions for photonic neural networks. Specifically, we investigate the Gaussian Error Linear Unit (GELU) \cite{hendrycks_gaussian_2023} and Sigmoid Linear Unit (SiLU) \cite{elfwing_sigmoid-weighted_2017} as alternatives to ReLU. These functions offer smooth, continuous derivatives across their entire domain, potentially providing more robust gradient propagation in noisy analog environments.

Unlike the sigmoid function, which is prone to the vanishing gradient problem during backpropagation \cite{pascanu_difficulty_2013}, GELU and SiLU are continuously differentiable variants of the rectified linear unit (ReLU) that may propagate gradients more effectively throughout deep neural networks. Recent work has demonstrated substantially higher accuracy for continuous activations such as GELU and SiLU compared to traditionally used discontinuous activations such as ReLU and LeakyReLU when noise is present \cite{xu_empirical_2015}. For example, in image classification, Shah et al. showed that GELU/SiLU equipped models were able to converge on simulated analog hardware, even in the presence of significant noise from multiple sources \cite{shah_analogvnn_2023}. Meanwhile, the ReLU equipped models struggled to show similar robustness to their GELU/SiLU counterparts. However, an explanation on the reasons underpinning this performance gap was not fully explored at the time. 

\begin{figure*}[t]
  \includegraphics[width=\textwidth]{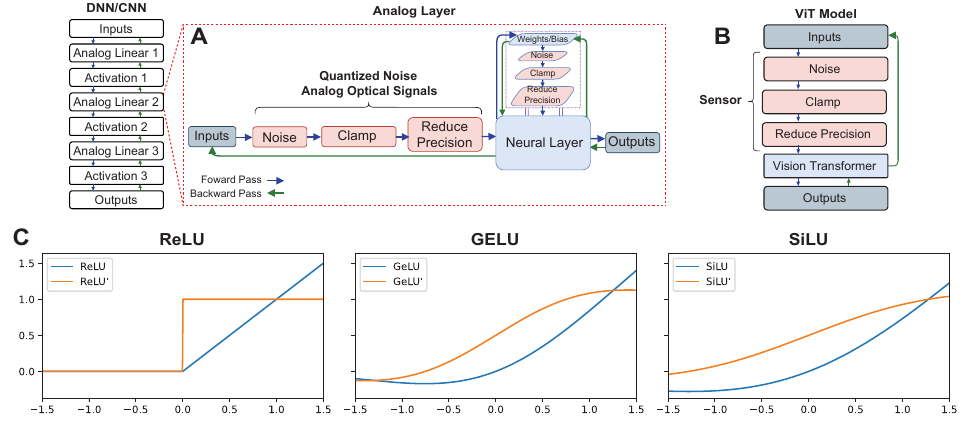}
  \caption{\textbf{Overview of model architecture.} \textbf{a)}  For the Linear, Convolutional, VGG and ResNet models we assume the worst case scenario where both the sensor and the model are physical and exhibit analog noise. For instance, this is the case for a CCD exhibiting electronic noise in conjunction with an analog photonic networks for computation. This is done by adding quantized noise layers between each traditional layer of the model. Adapted from \cite{shah_analogvnn_2023}. \textbf{b)} For a Vision Transformer model, the sensor could be implemented in analog hardware while the transformer network is implemented in digital hardware. \textbf{c)} ReLU, GELU and SiLU activation function and its derivatives}
  \label{fig:activation1}
\end{figure*}

In this work, we provide an explanation of why GELU/SiLU equipped models excel given noisy and quantized data such as the low-precision, quantized data provided by analog sensors or passed between layers of an optical neural network. We directly observe and quantify how the discontinuity in the derivatives of the ReLU activation function leads to error amplification during backpropagation under noise. In contrast, GELU/SiLU's continuous derivatives maintain stability and uniform backpropagation errors in the presence of noise (more information is provided in Section \ref{FunctionalAnalysis}). To understand how these continuously differentiable activation functions work with respect to analog noise sources across different neural network architectures, we also provide an analysis of the results of linear, convolution, VGG, ResNet, and transformer models, as shown in Section \ref{ExperimentalResults}. Overall, this work shows the superiority of GELU/SiLU in enabling more reliable, noise-resilient perception, prediction, and planning systems. Our findings provide guidance to analog system architects on selecting noise-resilient activations in real-world and real-time environments.

%% file: sections/2-Background.tex
\section{Background}
\subsection{Activations}
Neural networks rely on activation functions to introduce non-linearities that enable modeling complex patterns in data. The rectified linear unit (ReLU) activation function and its derivative are defined as:

\begin{equation}
\textrm{ReLU}(x) = \max(0, x)
\end{equation}
\begin{equation}
\textrm{ReLU}'(x) = \left\{ \begin{array}{cl}
1 & : \ x > 0 \\
0 & : \ x \leq 0
\end{array} \right.
\end{equation}

ReLU has been widely adopted due to its simplicity and effectiveness \cite{nair_rectified_2010}. However, ReLU has a discontinuity in its derivative at $x=0$ that can impede gradient flow and model training. ReLU neurons can also become stuck in a permanently deactivated state, known as the dying ReLU problem, hindering model expressiveness over time \cite{lu_dying_2020}.

Variants like LeakyReLU give a small negative slope instead of zero for $x < 0$ to mitigate this \cite{xu_empirical_2015}, but LeakyReLU still has a derivative discontinuity that may limit noise resilience. In contrast, GELU and SiLU are continuous variants of ReLU aimed at improving gradient flow.

GELU and its derivative are defined as:
\begin{equation}
\begin{split}
\textrm{GELU}(x) &= x\ \Phi(x) = x \cdot \frac{1}{2}\left[1 + \textrm{erf}\left(\frac{x}{\sqrt{2}}\right)\right] \\ 
\textrm{GELU}'(x) &= \frac{1}{2} \left(1 + \textrm{erf}\left(\frac{x}{\sqrt{2}}\right)\right) + \frac{x}{\sqrt{2 \pi}} e^{-\frac{x^2}{2}}
\end{split}
\end{equation}
where $\textrm{erf}(x)$ is the Gaussian error function. SiLU uses the logistic sigmoid instead and its derivative is defined as:
\begin{equation}
\begin{split}
\textrm{SiLU}(x) &= x\ \sigma(x) = \frac{x}{1+e^{-x}} \\
\textrm{SiLU}'(x) &= \frac{1 + e^{-x} + x e^{-x}}{\left(1 + e^{-x}\right)^2} \\
\end{split}
\end{equation}

As shown in Figure \ref{fig:activation1}C, both GELU and SiLU maintain continuity in their function and derivative, suggesting more stable gradients. Recent work has shown improvements in accuracy from using these activations, particularly in noise-corrupted settings \cite{shah_analogvnn_2023}. 

In the context of vision transformers, an additional variant known as GeGLU (Gaussian Error Gated Linear Unit) and ReGLU (Rectified Gated Linear Unit) are explored \cite{shazeer_glu_2020}. These transformer-unique activations are defined as:

\begin{equation}
\begin{split}
\textrm{ReGLU}(x, W, V, b, c) & = \textrm{max}(0, xW + b) \otimes (xV + c) \\
\textrm{GeGLU}(x, W, V, b, c) & = \textrm{GELU}(xW + b) \otimes (xV + c) \\
\end{split}
\end{equation}

\subsection{Analog/Photonics Errors}

All analog or real-life devices like cameras, sensors, and photodetectors ultimately employ analog-to-digital conversion to transform continuous analog signals into discrete digital data. However, real-world analog signals intrinsically suffer from noise. When physical implementations, such as photonic, neuromorphic, or quantum systems, are used to implement neural network models, this omnipresent noise permeates both the inputs as well as the inter-layer signals due to the analog-to-digital conversion process. This type of quantization noise can be simulated by adding a Gaussian Noise Layer, a Reduced Precision Layer, and a Clamp Layer (as shown in Figure \ref{fig:activation1}) \cite{shah_analogvnn_2023}. Here we define three parameters used to evaluate our models in the presence of low-precision and analog noise:

\paragraph{Error Probability (EP)} This is the probability that the recorded digital value is different from the true analog signal due to the presence of noise. That is, it is the probability that a reduced precision analog data point acquires a different digital value after passing through both a noise layer and then a reduced precision layer. The relationship between EP, photodetector/sensor bit precision (\(b\)), and standard deviation (\(\sigma\)) is defined as follows: 
\begin{equation}
    \textrm{EP} = 1 - \textrm{erf} \left( \frac{1}{2\sqrt{2}\ \sigma\ \left( 2^{b} - 1 \right)} \right)
\end{equation}

\paragraph{Reduced Precision (RP)} Reduce Precision layer applies a round-to-nearest transformation to the input based on the precision (number of discrete levels) \cite{shah_analogvnn_2023}.
\begin{equation}
    \textrm{RP}(x) = \frac{1}{2^p} sign(x) \left\lceil \left| 2^p \cdot x \right| - 0.5 \right\rceil
\end{equation}
where $p$ is the bit-precision.

\paragraph{Gradient Step Discontinuity (GSD)} This is the size of the step discontinuity present in the derivative of an activation function. For example, the gradient step discontinuity for ReLU at zero is 1 and for GELU/SiLU at zero is 0. 
\begin{equation}
    \textrm{GSD}_{f}(x_0) = \left| \lim_{x \to x_0^-} f'(x) - \lim_{x \to x_0^+} f'(x) \right| 
\end{equation}

%% file: sections/3-FunctionalAnalysis.tex
\section{Methods}
\subsection{Model Architecture}
The model architectures used in this work are illustrated in Figure \ref{fig:activation1}. The analog implementation of convolutional neural network (ConvNet, 6 convolutional layers + 3 linear layers) \cite{shah_analogvnn_2023}, VGG-A \cite{simonyan_very_2015} and ResNet-18 \cite{he_deep_2015} are shown in Figure \ref{fig:activation1}A. To simulate the worst-case scenario of both the photonic layers and input sensor data being subjected to noise and limited precision, we insert quantization noise layers between each typical linear or convolutional layer of the model architecture, as well as on the weights and biases within that layer. This represents a fully analog system implementation on an integrated analog hardware chip. In contrast, for the vision transformer presented in Figure \ref{fig:activation1}B, we assume only the photodetector/sensor data is analog, but the model is otherwise implemented digitally. Thus, quantized noise layers are added only to the inputs and not within the transformer network itself. The transformer network has a depth of 4 layers and 8 attention heads. 

Table \ref{table:hyperparameters-table} presents the hyperparameters subjected to experimentation across all models. In each experiment, the models underwent standard training and testing processes using Gaussian noise, Clamp normalization within the range of -1 to 1, and reduced precision, conducted on the CIFAR-10 and CIFAR-100 datasets.

\begin{table}[!h]
\vskip 0.15in
\begin{center}
\begin{small}
\begin{sc}
\begin{tabular}{lcccr}
\toprule
Hyperparameter & Parameters Tested \\
\midrule
Models    & \makecell{ConvNet, VGG-A, \& ResNet-18 \\} \\
Color    & True, False \\
Bit-Precision & 2, 3, 4, 5, 6 \\
 Error Probability & 0.2, 0.3, 0.4, 0.5, 0.6, 0.7, 0.8 \\
Normalization & Clamp(-1, 1) \\
Noise & Gaussian \\
Precision & ReducePrecision \\
Dataset & CIFAR-10, CIFAR-100 \\
\bottomrule
\end{tabular}
\end{sc}
\end{small}
\end{center}
\caption{List of hyperparameters tested for all models}
\label{table:hyperparameters-table}
\end{table}

\subsection{Functional Analysis}
\begin{figure*}[t]
  \includegraphics[width=\textwidth]{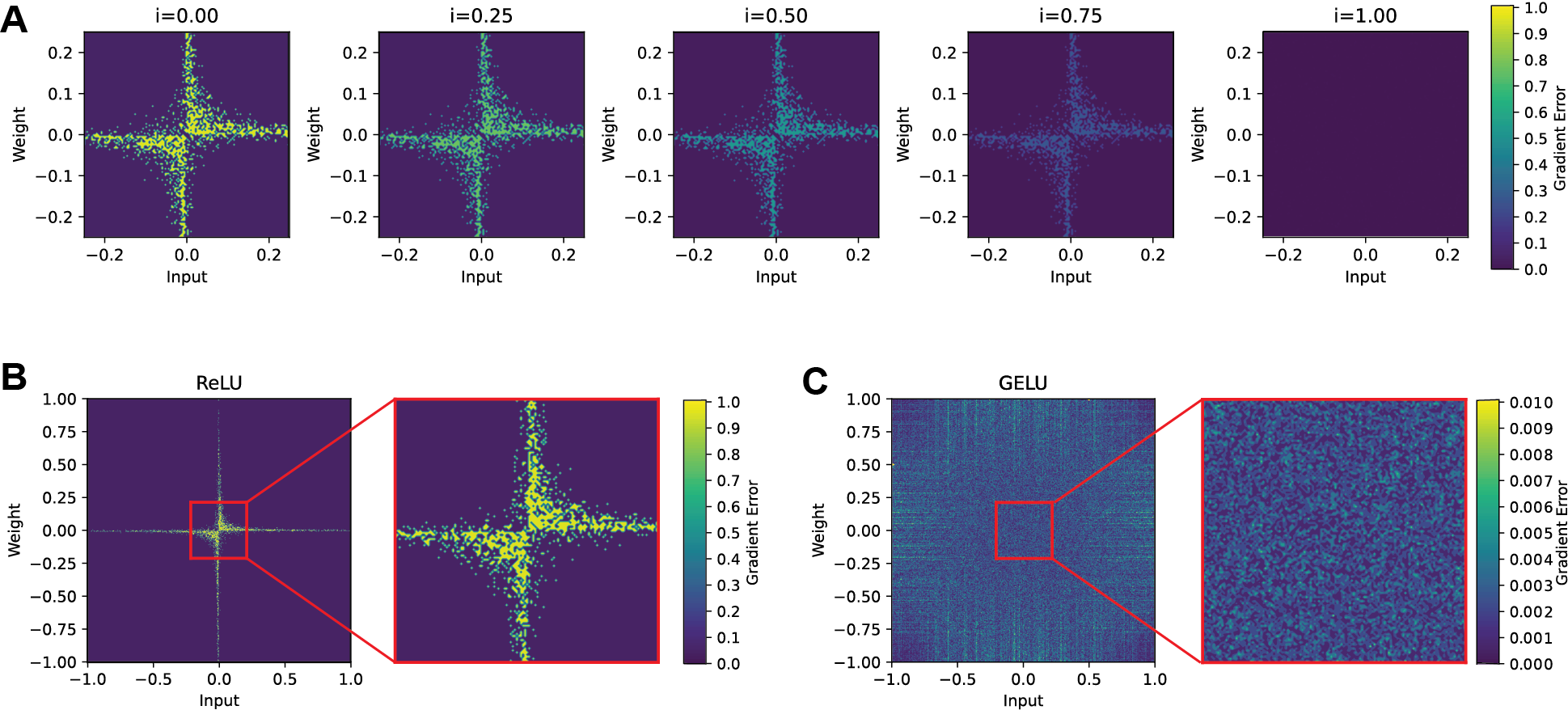}
  \caption{\textbf{Gradient Error in ReLU and GELU when inputs and weight are 8-bit quantized with 0.5 error probability for noise in a linear layer with no bias followed by an activation layer.} \textbf{a)} shows the gradient errors when interpolating between ReLU and GELU. Gradient Error \textbf{b)} for ReLU activation; \textbf{c)} for GELU activations;. \textbf{NOTE}: the gradient error colorbar axis is different for ReLU and GELU in (B) and (C) respectively.}
  \label{fig:ErrorAnalysis1}
\end{figure*}
\label{FunctionalAnalysis}
As photodetector/sensor data is inherently bounded within maximum and minimum values, normalization can rescale these signals to the range [-1, 1]. However, unlike ReLU, the Gaussian error linear unit (GELU) activation function does not approach zero for negative input values until approximately $-2.5$ and is limited to approximately $+0.84$ for a normalized maximum input value of $+1$. To understand the effects of change in the input domain over which GELU's response is non-zero, we introduce an adjustable scaling factor ($s$) that multiplies the input ($x$) in the error function ($\textrm{erf}(x)$). This scaling factor effectively controls the input domain over which the GELU response is not close to zero. An appropriate choice of the scaling factor may provide a mechanism to match the effective range of GELU activation to the normalized analog input signals. (as seen in Figure \ref{fig:FunctionalAnalysis1}A):
\begin{equation}
    \textrm{GELU}(x) = x \cdot \frac{1}{2} \left[ 1 + \textrm{erf}\left(\frac{s \cdot x}{\sqrt{2}}\right) \right]
\end{equation}
While increasing the scaling factor to $s=3$ reduces signal attenuation after activation, this comes at the cost of reduced effective precision near zero when the input signal is quantized as shown in Figures \ref{fig:FunctionalAnalysis1}B-C. This can negatively impact model accuracy when inputs with limited precision (e.g., 6-bits) are also affected by noise (see Figure \ref{fig:FunctionalAnalysis1}D).

For effective comparison between differentiable and non-differentiable activation functions, a linear functional interpolation was used (as seen in Figure \ref{fig:ExperimentalResults1}A):
\begin{equation}
\begin{split}
\textrm{I}_{\textrm{GELU}}(x) &= \textrm{ReLU}(x) + i (\textrm{GELU}(x) - \textrm{ReLU}(x)) \\
\textrm{I}_{\textrm{SiLU}}(x) &= \textrm{ReLU}(x) + i (\textrm{SiLU}(x) - \textrm{ReLU}(x)) \\
\textrm{I}_{\textrm{GeGLU}}(x) &= \textrm{ReGLU}(x) + i (\textrm{GeGLU}(x) - \textrm{ReGLU}(x))
\end{split}
\end{equation}
where $i \in [0, 1]$, for $i=0$, $I_X$ is ReLU/ReGLU and $i=1$, $I_X$ is X.

A key observation in Figure \ref{fig:ExperimentalResults1}A is that a discontinuity exists and is maximized in the derivative when the input value is zero and the interpolation factor is set to zero, which represents a standard rectified linear unit (ReLU) activation function. As the interpolation factor increases, the discontinuity in the derivative decreases until it completely disappears when the interpolation factor reaches 1. Hence, the interpolation factor has a negative correlation with the smoothness of the derivative around the zero input value. This is further proven using gradient step discontinuity for $\textrm{GSD}_{\text{I}_{\textrm{GELU}}}(0)$ and $\textrm{GSD}_{\text{I}_{\textrm{SiLU}}}(0)$ is equal to $1 - i$, where $i$ is the interpolation factor, as is shown in Figure \ref{fig:ExperimentalResults1}D, where we can clearly see that the interpolation factor is negatively correlated with the discontinuity in derivative of the ReLU-GELU interpolation function.

%% file: sections/4-ErrorAnalysis.tex
\section{Error Analysis}
When inputs are reduced in terms of precision, the effects of gradient step discontinuity around zero become even more apparent, as illustrated in Figure \ref{fig:ExperimentalResults1}B. In the presence of noise, gradients close to zero can become highly uncertain due to the inherent discontinuity caused by ReLU activations, as confirmed in the left subplot of Figure \ref{fig:ExperimentalResults1}C. This uncertainty is absent with continuously differentiable GELU activation, as evidenced in the right subplot of Figure \ref{fig:ExperimentalResults1}C. The discontinuities introduced by ReLU activations thus make models more sensitive to uncertainties caused by lower precision inputs near zero in the presence of noise. We further investigate this phenomenon in more detail below.

The equation for an activation function's input value after it passes through a linear layer with quantized noise in inputs and weights (but with no bias) can be written as follows:
\begin{multline}\label{eq:x_activation}
    x_{activation} = \frac{1}{p^2}\text{sign}(x_ix_w) \left\lceil |x_ip| - 0.5 \right\rceil\left\lceil |x_wp| - 0.5 \right\rceil\\ + \frac{\epsilon}{p}\sqrt{\left\lceil |x_i p| - 0.5 \right\rceil^2+ \left\lceil |x_w p| - 0.5 \right\rceil^2} + \epsilon^2
\end{multline}

where $x_i$ is the input value, $x_w$ is the weight, $p$ is precision ($p=2^{\text{bit-precision}}$), and $\epsilon$ is a Gaussian random variable with $\mu=0$. Note that in this case, we assume a single input value and weight for ease of illustration, but a similar analysis holds when the dot-product between input and weight vectors approaches zero.

Figure \ref{fig:ErrorAnalysis1}A shows the effects of Equation \ref{eq:x_activation} on the gradient calculation of the ReLU and GELU activation functions at various interpolation factors. From Figure \ref{fig:ErrorAnalysis1}A, it can also be seen that when inputs or weights are close to zero (which is not uncommon for weights), noise will dominate, causing errors in the gradients of the activation function. In the case of ReLU (Figure \ref{fig:ErrorAnalysis1}B), the error in gradients are $\sim$100$\times$ higher than those observed for GELU (Figure \ref{fig:ErrorAnalysis1}C) when both inputs and weights have quantized noise in a linear layer with no bias. Notably, the error in GELU models are much more uniformly distributed than the gradient errors reported in ReLU models. 

Neural network training typically employs mini-batch optimization, where gradients are accumulated across a subset of training samples before updating network weights. The accumulated error introduced due to the activation functions can be mathematically represented as demonstrated in Equation \ref{eq:AccumelateErrorActivation}.

\begin{equation}
\label{eq:AccumelateErrorActivation}
E_{f}(x, n) = \frac{\sum_{i=0}^{n} f(x + \epsilon_i)}{n}
\end{equation}
where $E$ represents the mean error introduced due to activation in a mini-batch, $f$ is the activation function, $n$ is the mini-batch size, and $\epsilon$ is a Gaussian random variable with mean $\mu = 0$ and a small standard deviation $\sigma$. As the mini-batch size ($n$) increases, we find:
\begin{equation}
\label{eq:AccumelateErrorActivation2}
\begin{split}
E_{ReLU'}(0^+, n) & \rightarrow 0.5 \neq 1 = ReLU'(0^+) \\
E_{ReLU'}(0^-, n) & \rightarrow 0.5 \neq 0 = ReLU'(0^-) \\
E_{GELU'}(0^+, n) & \rightarrow 0.5 = GELU'(0^+)\\
E_{GELU'}(0^-, n) & \rightarrow 0.5 = GELU'(0^-) \\
\end{split}
\end{equation}

As illustrated in Equation \ref{eq:AccumelateErrorActivation2}, due to mini-batch training, the gradients of both GELU and ReLU activation functions near zero converge to 0.5 when quantized input errors are present. However, critical differences emerge in error propagation. For GELU, the accumulated error closely matches its true gradient value of 0.5. In contrast, ReLU exhibits significant discrepancies, with its true gradient near zero alternating between 0 and 1, while the accumulated error consistently approaches 0.5.

Due to this asymmetry, the ReLU activation function systematically suppresses any positive or negative weight bias information because of its gradient discontinuity, as it never provides the true gradient value for positive or negative weights. However, GELU activation does provide the true gradient value after mini-batch training. It is well established that weight biases are critical to neural network learning, as they enable sophisticated information storage and nuanced feature extraction from complex datasets. \cite{wang_bias_2019, bolager_sampling_2023} Thus, the uniformity of gradient errors from continuous differentiable activations such as GELU facilitate more reliable convergence with increasing model complexity on analog platforms.

%% file: sections/5-Results.tex
\section{Results}\label{ExperimentalResults}

The results presented here comprehensively demonstrate the superior noise resilience of continuously differentiable activations like GELU and SiLU compared to the traditionally used discontinuous ReLU activation.

Firstly, Figures \ref{fig:ExperimentalResults1}E and \ref{fig:ExperimentalResults1}F clearly show that linearly interpolating between the differentiably discontinuous ReLU and continuous GELU/SiLU activations leads to substantial gains in accuracy as the interpolation factor is increased. That is, systematically reducing the gradient step discontinuity of the activation function through interpolation significantly enhances model test accuracy on noisy quantized inputs. This also holds true for the CIFAR-100 dataset and across different types of model architectures as shown in Figure \ref{fig:ExperimentalResultsCIFAR100}.

Figure \ref{fig:FunctionalAnalysis1}D shows that limiting the input range where GELU responds non-linearly causes a small drop in model accuracy. This accuracy reduction stems from compressing the GELU function, which effectively decreases the precision of gradients around zero as seen from Figure \ref{fig:FunctionalAnalysis1}B-C. The less precise gradients make small weight adjustments more difficult, slightly hindering model performance. However overall this indicates that GELU is inherently robust to the bounded normalized inputs typical for optical hardware such as laser power, photocurrent, and sensor data. 

Furthermore, Figures \ref{fig:ExperimentalResults2}A-D examine how these errors can cause failure in model convergence as the number of layers is incrementally increased. As expected, adding more convolutional and linear layers leads to compounding of analog noise effects since the quantized noise is added to both inputs and models (as shown in Figure \ref{fig:activation1}A). During training, this causes the inaccuracy of weight updates to increase in ReLU equipped models, but not in GELU/SiLU equipped models. Notably, in certain deeper configurations, the models with ReLU units are completely unable to learn until the interpolation factor is sufficiently high---meaning the gradient step discontinuity is small enough that noise no longer impedes gradient convergence.

To better understand the influence of differentiation on the activation function, we also evaluated the impact of negative slope on LeakyReLU for a sub-set of our listed hyperparameters and observed similar challenges in model convergence. For LeakyReLU with varying negative slopes ($\alpha$), as the negative slope increases the gradient step discontinuity decreases, effectively making the function more continuous. We observed that after gradient step discontinuity is below a certain threshold, the model begins to learn effectively (detailed results are provided in Table \ref{table:LeakyReLU}), this is similar to what we observed in Figure \ref{fig:ExperimentalResults1}E \& \ref{fig:ExperimentalResults1}F. This finding further underscores the critical role of differentiability in enhancing robustness against noise, strengthening the argument for using differentiable activation functions.

\begin{table}[!h]
    \centering
    \begin{tabular}{l|l}
        \textbf{Activations} & \textbf{Top-1 Accuracy (\%)} \\ 
        \hline
        \textbf{ReLU} & \cellcolor{red!25} 10.78 \\ 
        \textbf{LeakyReLU ($\alpha = 0.01$)} & 10.30 \\ 
        \textbf{LeakyReLU ($\alpha = 0.1$)} & 68.58 \\ 
        \textbf{LeakyReLU ($\alpha = 0.2$)} & 71.41 \\ 
        \textbf{LeakyReLU ($\alpha = 0.3$)} & 72.48 \\ 
        \textbf{LeakyReLU ($\alpha = 0.4$)} & 69.11 \\ 
        \textbf{GeLU} & \cellcolor{green!50} 77.57 \\ 
    \end{tabular}
    \vspace{2mm}
    \caption{Top-1 Classification Accuracy of ConvNet with ReLU, GeLU and LeakyReLU ($\alpha$ is the negative slope) on CIFAR-10 Dataset}
    \label{table:LeakyReLU}
\end{table}

Finally, the vision transformer (ViT) results in Figure \ref{fig:ExperimentalResults3} confirm the consistent benefits of continuous activations over discontinuous variants for handling noise and reduced precision input data. Figures \ref{fig:ExperimentalResults3}A and \ref{fig:ExperimentalResults3}B both show an increasing accuracy trend as the interpolation factor rises. However the increase in model accuracy is not as pronounced as it is for linear and convolutional models. This is because, in the case of the ViT, the quantized noise is only added to the inputs but not to the model weights (as shown in Figure \ref{fig:activation1}B). In Figure \ref{fig:ExperimentalResults3}C, the correlation between the interpolation factor and accuracy for ReGLU-GeGLU can not be seen. This can be because ReGLU and GeGLU are much more complex functions and their differentiabilities cannot be easily evaluated with interpolation alone while quantization noise is present. However, the overall maximum accuracy of the ViT is still higher when using GELU/SiLU in comparison to ReGLU and GeGLU, demonstrating the importance of a fully differentiable activation function when quantization noise is present.

In summary, both the interpolation analysis and model depth analysis substantiate that the differentiability of GELU/SiLU activations enable superior gradient flow and noise resilience as compared to ReLU alternatives across various model architectures. The findings provide clear guidance for selecting activations to mitigate the impacts of unavoidable noise in real-world analog systems.

%% file: sections/6-Discussion.tex
\section{Discussion} \label{Discussion}

This work expands our understanding of differentiable activation functions' effectiveness and applicability in real-world quantized noise scenarios, building upon established knowledge of their advantages over non-differentiable activation functions. We demonstrate this benefit concretely for low-precision, high-noise settings that are common in real-world analog systems like photonic accelerators, memristive crossbar arrays, and other analog hardware. At error probabilities as low as 30\%, many model configurations with ReLU already struggle to learn, as shown in Figures \ref{fig:ExperimentalResults1}E and \ref{fig:ExperimentalResults1}F. In contrast, GELU/SiLU models are able to learn even at higher noise levels. Remarkably, we never observe any cases where ReLU outperforms the continuous activations such as GELU/SiLU. This suggests that the continuity principles we have identified may have broad applicability for enhancing model robustness.

More broadly, our findings also aid the explainability and interpretability of deep learning model training in real-world noisy environments, shedding light on how noise can affect model convergence. With neural networks remaining largely black boxes, understanding why certain architectures or components perform better guides debugging and continued progress. By pinpointing derivative continuity as the differentiating factor, we can directly advise system architects to utilize smooth activations to mitigate the effects of unavoidable noise sources in emerging analog accelerators. More generally, we advise those who work with such systems to examine all the non-linear model components which may propagate noise in a similar fashion.  

The implications are particularly important for emerging analog computing platforms like photonic, neuromorphic or quantum systems, which promise massive speed and power efficiency improvements but suffer from intrinsic hardware non-idealities. Our analysis indicates that realizing the performance potential of analog accelerators requires joint optimization of both the hardware and the model architecture. Additionally, the continuity principles established here may also aid in adversarial and out-of-distribution robustness in other applications, which warrants further investigation. Furthermore, in robotics and autonomous vehicles, where deep learning models process sensor data, this study helps not only with model training but also enables the creation of robust models providing more consistent outputs amidst noise. This enhances decision-making capabilities in dynamic, safety-critical scenarios.  

Finally, these findings could enable systems that use reduced sensor precision for low-power requirements to maintain or improve real-world performance. Such findings could be particularly interesting for reduced precision neural networks implemented on FPGAs. Typically reducing precision helps use far less silicon area and also enables faster inference (small bit precision systems can often employ fixed-point or integer arithmetic which is faster than floating-point counterparts). This is analogous to the increased speed and robustness seen with 4-level PAM modulation over binary implementations \cite{zhang_resource-constrained_2022, ngadiuba_compressing_2020, wielgosz_mapping_2019}. By reducing precision, neural network inferences can operate faster while using less space and power---crucial metrics for embedded applications like robotics and autonomous vehicles.

%% file: sections/7-Conclusion.tex
\section{Conclusion}
In this work, we have demonstrated the noise resilience advantages of continuously differentiable activations over discontinuous rectified activations through extensive functional analysis and model training across a range of neural network architectures. Our investigations conclusively establish that the built-in continuity of GELU and SiLU derivatives enable reliable gradient flows that mitigate the impacts of errors arising from common analog noise sources such as quantization and hardware non-idealities.

The key findings can be summarized as follows:

\begin{itemize}
    \item GELU and SiLU exhibit inherent robustness to inputs bounded within normalized photodetector/sensor ranges. In contrast, ReLU suffers from uncertainty in gradients that leads to unstable convergence and amplification of errors during backpropagation.
    \item Interpolating between discontinuous ReLU and continuously differentiable activations systematically improves accuracy as the gradient step discontinuity decreases. This establishes a definitive causal link between activation gradient step discontinuity and noise resilience.
    \item Continuity advantages accumulate with model depth, leading to larger improvements from GELU/SiLU usage in deeper networks where analog errors would otherwise compound from discontinuous activations.
    \item Vision transformer results corroborate the consistent benefits of differentiable activations with their convolutional and linear architectured counterparts - affirming the importance of intrinsic activation gradient continuity on modern deep-learning architectures.
\end{itemize}

Together, these comprehensive results and analyses provide guidance to hardware-based model architects to employ smooth activations such as GELU and SiLU in order to fully realize the performance potential of emerging analog platforms, such as photonic accelerators. By selecting appropriate activations, the detrimental impacts of real-world noise sources can be significantly reduced without requiring excessive precision requirements. While our focus has been on robustness to sensor and analog hardware-induced noise, the continuity principles established here may also translate to improved stability for other applications. These may include adversarial robustness and out-of-distribution detection, which might warrant future investigation \cite{liang_detecting_2021, yang_generalized_2024}. Overall, by explaining and demonstrating the underlying mechanisms relating continuity to resilience, this work helps pave the path toward reliable, performant, and fully analog AI implementations.

%% file: sections/10-Appendix.tex
{
\appendix[Effects of Learning Rate]
\label{section:LearningRate}
We find that even over a wide range of learning rates, only functions with small discontinuities are able to successfully learn features from the training dataset. Table \ref{table:LearningRate} illustrates this using ConvNet trained on the CIFAR-10 dataset.

\begin{table}[!h]\label{table:LearningRate}
    \centering
    \begin{tabular}{l|llll}
        \textbf{} & \textbf{0.1} & \textbf{0.01} & \textbf{0.001} & \textbf{0.0001} \\ 
        \hline
        \textbf{0 (ReLU)} & 10.01 & 10.10 & 10.53 & 10.65 \\ 
        \textbf{0.1} & 10.02 & 10.13 & 10.83 & 10.55 \\ 
        \textbf{0.2} & 10.04 & 10.1 & 10.69 & 11.02 \\ 
        \textbf{0.3} & 10.17 & 10.22 & 10.61 & 10.8 \\ 
        \textbf{0.4} & 10.38 & 10.54 & 10.48 & 10.7 \\ 
        \textbf{0.5} & 10.50 & 10.05 & 10.32 & 10.75 \\ 
        \textbf{0.6} & 10.10 & 10.21 & 10.58 & 10.85 \\ 
        \textbf{0.7} & 10.01 & 10.17 & \cellcolor{green!50}69.11 & \cellcolor{green!15}57.31 \\ 
        \textbf{0.8} & 10.66 & 10.15 & \cellcolor{green!50}70.28 & \cellcolor{green!15}59.72 \\ 
        \textbf{0.9} & 10.15 & 10.18 & \cellcolor{green!50}71.71 & \cellcolor{green!15}59.24 \\ 
        \textbf{1 (GELU)} & 10.29 & 10.29 & \cellcolor{green!50}71.55 & \cellcolor{green!15}58.13 \\ 
    \end{tabular}
    \vspace{2mm}
    \caption{Top-1 Accuracy for Interpolation Factor (rows) vs Learning rate (columns) on ConvNet with CIFAR-10 at 0.8 error probability}
\end{table}
}